\documentclass[10pt,twocolumn]{article} 
\usepackage{simpleConference}
\usepackage{times}
\usepackage{graphicx}
\usepackage{amssymb}
\usepackage{url,hyperref}
\usepackage{comment}

\begin{document}

\title{How to Choose a Threshold for an Evaluation Metric for Large Language Models}

%\author{Iroro Orife \\
%\\
%Technical Report \\
%Seattle, Washington, USA \\
%\today
%\\
%\\
%iroro@alumni.cmu.edu  \\
%}

\author{Bhaskarjit Sarmah\\
    BlackRock, Inc.\\
	Gurgaon, HR, India \\
	\texttt{bhaskarjit.sarmah@blackrock.com}\\
        \and
        Mingshu Li \\
	BlackRock, Inc.\\
	Atlanta, GA, USA\\
	\texttt{mingshu.li@blackrock.com} \\
	%% examples of more authors
	\and
	{Jingrao Lyu} \\
	BlackRock, Inc.\\
	Atlanta, GA, USA \\
	\texttt{jingrao.lyu@blackrock.com} \\
	\and
	{Sebastian Frank}\\
    BlackRock, Inc.\\
	New York, NY, USA \\
	\texttt{sebastian.frank@blackrock.com} \\
 \and
 {Nathalia Castellanos}\\
    BlackRock, Inc.\\
	Atlanta, GA, USA \\
	\texttt{nathalia.castellanos@blackrock.com} \\
 \and
	{Stefano Pasquali} \\
	BlackRock, Inc.\\
	New York, NY, USA \\
	\texttt{stefano.pasquali@blackrock.com} \\
 \and
	{Dhagash Mehta} \\
	BlackRock, Inc.\\
	New York, NY, USA \\
	\texttt{dhagash.mehta@blackrock.com} \\
    }

\maketitle
\thispagestyle{empty}

\begin{abstract}
To ensure and monitor large language models (LLMs) reliably, various evaluation metrics have been proposed in the literature. However, there is little research on prescribing a methodology to identify a robust threshold on these metrics even though there are many serious implications of an incorrect choice of the thresholds during deployment of the LLMs. Translating the traditional model risk management (MRM) guidelines within regulated industries such as the financial industry, we propose a step-by-step recipe for picking a threshold for a given LLM evaluation metric. We emphasize that such a methodology should start with identifying the risks of the LLM application under consideration and risk tolerance of the stakeholders. We then propose concrete and statistically rigorous procedures to determine a threshold for the given LLM evaluation metric using available ground-truth data. As a concrete example to demonstrate the proposed methodology at work, we employ it on the Faithfulness metric, as implemented in various publicly available libraries, using the publicly available HaluBench dataset. We also lay a foundation for creating systematic approaches to select thresholds, not only for LLMs but for any GenAI applications.
\end{abstract}

\section{Introduction}
Transformer architectures \cite{vaswani2017attention} and then subsequent developments in Large Language Models (LLMs) \cite{zhao2023survey} have become one of the most important inventions in the field of artificial intelligence (AI) in recent times that enables machines to understand and generate \textit{human-like} text. LLMs have found a plethora of applications across various sectors, including healthcare, education, finance, and entertainment. From customer-facing virtual assistants, chatbots, and information extraction from documents to automated content creation and language translation, LLMs are increasingly being integrated into technologies that will potentially interact with millions of users on a regular basis. A widespread adoption of LLMs also underlines the importance of ensuring that these models operate reliably, responsibly, and ethically.

To validate and continuously monitor the reliability (in various senses) of LLMs, a variety of evaluation metrics have been proposed in the literature \cite{guo2023evaluating}. These metrics can be broadly categorized into reference-based and reference-free evaluations. Reference-based metrics, such as BLEU (Bilingual Evaluation Understudy) \cite{papineni2002bleu}, ROUGE (Recall-Oriented Understudy for Gisting Evaluation) \cite{lin2004rouge}, and BERTScore \cite{zhang2019bertscore}, compare the text generated by the given model against ground truth reference text to measure accuracy, fluency, and semantic similarity. Reference-free metrics evaluate the quality of generated text without relying on any reference ground truth text. For example, perplexity evaluates how well the model predicts the next word in a sequence, indicating fluency and coherence based on the model's own probability distributions. Faithfulness, Answer Relevance, etc.\cite{es2023ragas, guo2023evaluating} metrics quantify how accurately a generated response reflects the retrieved context and how relevant the response is to the given query, respectively, for a retrieval augmented generation (RAG) system \cite{lewis2020retrieval}. Often, reference-free evaluation metrics either involve human judgments or another LLM. Both types of metrics are essential for a comprehensive evaluation of LLMs, capturing different aspects of an LLM's performance that may be critical for real-world applications.

Unlike traditional evaluation metrics for models for tabular data, such as $R^2$, root mean squared error, precision, recall, etc., many of the LLM evaluation metrics rely on another (sometimes blackbox) model themselves. Moreover, they usually have a continuous range of values, though some of them may have only finitely many possible values such as 'hallucinated' and 'not hallucinated'. Selecting the correct threshold for these evaluation metrics with a continuous allowed range is not only crucial for technical accuracy but also has significant societal implications because the thresholds determine when the output of the given model is acceptable. Often, the thresholds on the evaluation metrics may solely influence whether the generated responses meet the necessary standards for reliability and safety. In mission-critical systems in domains like aviation, healthcare, law, or finance, an inappropriate threshold can lead to the dissemination of misinformation, potentially causing harm to individuals and communities.

The thresholds may also play a vital role in shaping the societal impact of AI by balancing sensitivity and specificity. An overly lenient threshold may result in a high number of false positives, allowing harmful or biased content to spread unchecked. This can perpetuate stereotypes, reinforce social inequalities and biases, financial instabilities, or incite tension and conflict. On the other hand, an excessively strict threshold can suppress valuable information, hindering innovation, restricting access to beneficial knowledge, or can simply make the downstream application useless for the end users. By carefully calibrating thresholds, LLM developers can mitigate these risks.

Despite the critical importance of selecting appropriate thresholds for evaluation metrics in LLMs, while significant effort has been devoted to developing various metrics, the methodologies for setting optimal thresholds for these metrics remain underexplored. This lack of guidance leaves practitioners without clear strategies to balance factors such as precision and recall or to align model performance with specific application requirements and risk appetite. Consequently, this gap hampers the ability to fully optimize LLM performance and ensure their reliable deployment in real-world scenarios where threshold selection is pivotal.

In Ref.~\cite{Sudjianto2024}, the authors lay down a broader model risk management (MRM) approach for RAG systems, propose different ways to compute the context relevancy, groundedness, completeness and answer relevancy based on the text similarity metrics. They also propose a specific way to identify thresholds for LLM evaluation metrics. Although the paper comes close to the present work, there are various differences between the two: our work is focused solely on exploring different methods to identify thresholds for LLM evaluation metrics rather than the general MRM framework for LLMs of Ref.~\cite{Sudjianto2024}. Here, we propose to start the process of identifying the threshold from a behavioral finance point of view by first identifying the AI risk tolerance of the stakeholders, whereas Ref.~\cite{Sudjianto2024} focuses on calibrating the evaluation metrics results with human labels to ensure that the scores align with human perceptions. 

In the present work, we provide a list of different statistical methods to carefully select particular thresholds and demonstrate how to utilize some of them in concrete examples using publicly available datasets and libraries for the LLM evaluation metrics for reproducibility purposes. In short, the present work is focused on more concrete prescription on methodologies and the actual computation to pick thresholds for the LLM evaluation metrics in a statistically rigorous way rather than the end-to-end MRM for a RAG system, and hence should be viewed as complementary work to Ref.~\cite{Sudjianto2024}. 

In the remainder of the paper we begin by laying down a detailed description of the methodology to compute a threshold for the LLM evaluation metrics. Then, we focus on a publicly available dataset called HaluBench \cite{ravi2024lynx}, which is one of the rare datasets that contains around 15K question, context, and answer triplets along with human annotated labels for hallucination, and apply the proposed techniques to identify thresholds on a popular LLM evaluation metric called Faithfulness. After discussing the results and their implications, we provide an outlook and conclusion.

\section{Proposed Methodology to Determine Threshold on LLM Evaluation Metrics}\label{sec:proposed_method}

We propose a systematic approach to determine a threshold for an LLM evaluation metric having a continuous allowed range. The proposed approach is based on the conventional MRM practices in the financial industry \cite{reserve2011supervisory,sudjianto2024model} as well as National Institute for Standards and Technology (NIST) AI Risk Management Framework Playbook \cite{ai2023artificial}. The following are the steps we propose:

\subsection{Step 1: Identify and Quantify Risks of the Specific Use Case and Risk Tolerance of the Stakeholders}
Before discussing any statistical methodology to analyze the given LLM evaluation metric and then pick a threshold, we propose to start with identifying the risk tolerance measured from different dimensions. There are at least two broad aspects we have to consider: the risks of the specific application when deployed, and the risk appetite of the business that owns the application.

\subsubsection{Understanding the risks of the specific application when deployed}
The MRM best practices yield to start by defining the LLM use case clearly, and specifying where and how the LLM will be deployed (e.g., internal to the business or external, chatbot, or other user-interface, etc.).

Identifying and enumerating potential risks for the potentially deployed LLM application and considering the consequences of possible negative outcomes of incorrect or suboptimal model outputs are also important follow-ups to the definition of the LLM use case. More specifically, it is important to identify all the potential legal, financial, reputational, regulatory, societal, customer experience related, etc. risks for the application to determine a holistic view on the risk profile of the LLM application under consideration. 

Additionally, since evaluating many of the commercially available LLMs itself may incur cost (e.g., cloud infrastructure, as well as in the case of LLM as a Judge, there may be additional LLM cost per token), and hence the stakeholders may also need to take the cost-benefit trade-off into account into the risk appetite calculations.

Note that each of these types or even granular sub-types of risks may be quantified, if they can be quantified that is, with the help of one or more evaluation metrics, and the modeler should choose the appropriate evaluation metric carefully.

Prescribing the precise risk rating methodology or which LLM evaluation metric should be chosen to mitigate what type of risk is beyond the scope of the present work, but will be discussed in a future work.

\subsubsection{Identify the Risk Tolerance of the Stakeholders}
It is also important to identify the stakeholders' underlying attitudes toward risk and their risk appetite towards the aforementioned risks of the specific LLM application when deployed. There is limited literature on identifying an individual's or an organization's (or even a whole country's) \textit{AI risk tolerance} that we loosely define as the extent to which a person is willing to accept and navigate potential risks associated with AI in decision-making for a given application. An obvious direction may be to borrow concepts from behavioral finance to model their risk preferences mathematically or at least heuristically. i.e., translating the Prospect Theory \cite{kahneman2013prospect} concepts that begin by recognizing that stakeholders may value losses more heavily than gains, affecting their tolerance for errors; then, presenting the stakeholders with a few hypothetical scenarios involving trade-offs between model accuracy and potential errors, fit their responses to a predetermined utility functions (e.g., exponential or logarithmic) to mathematically represent their risk aversion or risk-seeking behavior; and, eventually to extract the degree of risk aversion from the fit. 

Alternatively, one can also come up with a more simplistic methodology which requires the stakeholders to fill out a well-crafted questionnaires whose answers could eventually identify the individual's risk tolerance, though recent studies have revealed various limitations of this approaches \cite{thompson2021know,thompson2022measuring}.

Again, prescribing a specific methodology to identify the AI risk tolerance of the stakeholders is beyond the scope of the present work. 

\subsubsection{Translating the Risk Tolerance to a Statistical Quantity}
For the present purposes, the final goal of this exercise should be to provide an answer to a practical question such as what percentage of Type I (false positive) and Type II (false negative) errors the stakeholder's risk preference corresponds to for the chosen metric. In the below, we assume that we have this specific information from the stakeholders. 

To feed a concrete statistic into the downstream computation of the evaluation metric threshold, the risk preferences should be translated into a corresponding statistical confidence level (e.g., only 5\% hallucination is accepted for a specific application for moderate risk tolerance, and hence the required confidence level is 95\%) while ensuring that the confidence level reflects both practical considerations and psychological risk preferences.
    
\subsection{Step 2: Prepare Ground Truth Data}
Even though a given evaluation metric is reference-free (e.g., Faithfulness), validation of the evaluation metric and computation of its threshold require ground truth data. Here, the ground truth data can be potential questions (as diversified as possible so that they mimic the real-world scenario as closely as possible, see, e.g., \cite{Sudjianto2024} for an attempt to diversify the training data in a specific sense), the ground truth context, answer, and label corresponding to the chosen evaluation metric (such as 'hallucinated' or 'not hallucinated', for the Faithfulness score). Here, the data can be generated either manually or synthetically, but the labels may be human curated.

\subsection{Step 3: Determine the Threshold for the Metric and Cross-Validate}
Given the ground truth labeled data, partition the data into training and testing splits, compute the LLM evaluation metric under consideration for all of the samples in the training dataset, compute the threshold using one of the below statistical methods for the predetermined confidence level, and check if the threshold indeed provides expected results on the test data.

To ensure further robustness of the threshold, one should perform a more rigorous cross-validation (e.g., K-fold cross-validation, assuming that there is no temporal dependency among the internal documents).

Below is a sample list of quantitative methods to pick the threshold:

\subsubsection{Threshold using Z-score}
An intuitive approach for deriving thresholds at different confidence levels is to utilize the Z-score, which quantifies how many standard deviations a value lies from the mean under the assumption of a normal distribution \cite{hastie2009elements}. For example, under a standard normal distribution assumption, $P(-1.96<Z<1.96) = 0.95$, indicating a 95\% probability that a standard normal variable $Z$ will fall within this range. In the context of an evaluation metric for LLMs, this methodology can help translate machine-generated scores into actionable thresholds at user-specified confidence levels $1-\alpha$, using the following formula:
\begin{equation}\label{eq:z-score}
\bar{X}\pm{Z_{\frac{\alpha}{2}}\times{\frac{\sigma}{\sqrt{n}}}},
\end{equation}
where $\bar{X}$ is the sample mean of the evaluation metric, $\sigma$ is the population standard deviation, and $n$ is the sample size. This method provides a straightforward way to compute thresholds, but it relies on key assumptions. The Z-score approach is valid when the data follows a normal distribution or when the sample size is sufficiently large for the Central Limit Theorem to ensure approximate normality.

Though this method is the easiest to implement and interpret and does not even require the ground truth labels to derive the threshold (as opposed to verify the threshold), for certain LLM evaluation metrics such as faithfulness, the normal distribution assumption may obviously not hold as the values may be expected to concentrate near, say, 0 (e.g., hallucinated) or 1 (not hallucinated).

\subsubsection{Threshold using Kernel Density Estimation}

Determining a threshold for bimodal (or multimodal) distributed scores involves identifying the midpoint that effectively separates the two modes. One naive method estimates the density of the faithfulness score using a histogram or kernel density estimation (KDE) \cite{chen2017tutorial}. By identifying the two peaks (modes) in the distribution, the midpoint is located at the local minimum between these peaks. However, associating this midpoint with specific confidence levels proves challenging, making the method less flexible for threshold identification.

To address this limitation, we can leverage Bayes' rule on the estimated distribution of the faithfulness score for each label. Specifically, for any given faithfulness score $x$, the posterior probabilities of hallucination ($L_{0}$) or non-hallucination ($L_{1}$) are computed using Bayes' rule:
\begin{equation}\label{eq:kde}
P(L_{i}|x) = \frac{P(x|L_{i})P(L_{i})}{P(x)},
\end{equation}
where $P(x|L_{i})$ is the likelihood of $x$ given hallucination label $L_{i}$, $P(L_{i})$ is the prior probability of label $L_{i}$, and $P(x)$ is the normalizing constant $P(x) = P(x|L_{1})P(L_{1}) + P(x|L_{0})P(L_{0})$.This approach enables the identification of thresholds corresponding to different confidence levels by examining their conditional probabilities, offering a probabilistic framework for threshold determination.

\subsubsection{Threshold using Empirical Recall}
Another approach involves examining the relationship between empirical recall and the faithfulness score. The faithfulness scores are first sorted in ascending order, with instances having higher scores classified as passing the hallucination test. For each threshold, recall (or any other evaluation metric of interest) is computed based on the ground-truth labels in the training dataset. This process is systematically repeated across the entire range of faithfulness scores to derive the empirical recall curve as a function of varying thresholds.

\subsubsection{Threshold using AUC-ROC}
Receiver Operating Characteristic (ROC) curves are commonly used to represent the performance of a binary or multi-class classifier across all possible probability thresholds \cite{hastie2009elements}. These curves plot the true positive rate (sensitivity) against the false positive rate (1 - specificity, or Type I error), illustrating the trade-offs associated with various cutoff values. A key summary metric derived from the ROC curve is the area under the curve (AUC) which reflects the classifier's ability to correctly distinguish which of two samples is more likely to be in one of the classes over the other(s).

To derive thresholds from the ROC curve, first, train a classifier such as logistic regression (binary or multi-class, as appropriate), decision tree, support vector classifiers, Random Forests, etc. with the evaluation score as the input and the ground truth labels as the target. Then, translate the risk tolerance into acceptable levels of false positives (Type I error) and false negatives (Type II error), and then identify the corresponding thresholds with the help of the ROC curve. For example, to achieve a 95\% confidence level, the threshold can be set at the point corresponding to a 5\% false positive rate on the ROC curve.

\subsubsection{Threshold using Conformal Prediction}
Conformal prediction is a model-agnostic framework for generating prediction intervals at specified confidence levels, relying only on the assumption of data exchangeability \cite{angelopoulos2023conformal}, it can be a valuable tool for quantifying uncertainty by relating evaluation measures to ground truth labels. This approach produces prediction sets that include the true label while catering to the user's specified risk tolerance. In this study, we employ split conformal prediction as an example, which uses a hold-out dataset to calibrate the empirical distribution of prediction errors, leveraging access to ground truth labels. The detailed workflow is described as follows.

First, a classifier is trained on the training data $\mathcal{I}_{1}$ to predict the ground truth label based on the evaluation metric. This step effectively maps the raw evaluation metrics to a calibrated probability space, aligning them with the true likelihood of the ground truth labels \cite{boken2021appropriateness}. This calibration technique is commonly referred to as Platt scaling.

Here, we explore various classification models to serve as the underlying classifier. The first is standard logistic regression, which acts as a baseline model. Logistic regression generates a sigmoid-shaped probability curve, providing a straightforward implementation of Platt scaling.
\begin{equation}\label{eq:logistic-regression}
P(Y=1|X) = \frac{1}{1+e^{-(\beta_{0}+\beta_{1}X_{1}+\beta_{2}X_{2}+...+\beta_{p}X_{p})}},
\end{equation} 
where $\beta_{0}$ is the intercept term, $\beta_{1},\beta_{2},...,\beta_{p}$ are coefficients for the predictors $X_{1}, X_{2},...,X_{p}$. 
However, standard logistic regression assumes a linear relationship between the evaluation metric and the ground truth labels, which limits its ability to capture complex relationships in the data. Hence, we also examine polynomial logistic regression \cite{hastie2009elements} and Generalized Additive Models (GAMs) \cite{hastie2017generalized} with logistic regression as the link function.

Specifically, polynomial logistic regression extends standard logistic regression by incorporating higher-degree polynomial features to model complex, non-linear relationships. For multiple predictors, the model can include interactions and cross products of predictors.

\begin{comment}
For a single predictor $X$, the model is expressed as:
\begin{equation}\label{eq:poly-singlefeature}
P(Y=1|X) = \frac{1}{1+e^{-(\beta_{0}+\beta_{1}X+\beta_{2}X^{2}+\beta_{3}X^{3}+...+\beta_{k}X^{k})}},
\end{equation} 
where $X^{k}$ is the polynomial terms of $X$, and $k$ is the degree of the polynomial. For multiple predictors, the model can include interactions and cross products of predictors.

\begin{equation}\label{eq:poly-multiplefeatures}
P(Y=1|X) = \frac{1}{1+e^{-(\beta_{0}+\sum_{i=1}^{p}\beta_{i}X_{i}+\sum_{i\leq{j}}\beta_{ij}X_{i}X_{j}+...)}}.
\end{equation}
\end{comment}

The non-parametric GAM models the relationship as an additive function of smooth terms, without making strict assumptions about the shape of these relationships. This flexibility allows GAMs to effectively capture intricate patterns and non-linear dependencies in the data.

\begin{comment}
\begin{equation}\label{eq:gam}
g(E(Y)) = \beta_{0} +f_{1}(X_{1}) +f_{2}(X_{2}) +...+f_{p}(X_{p}),
\end{equation}
where $g$ represents the link function, and $f_{1},f_{2},...,f_{p}$ are smooth, non-linear functions of the evaluation metrics. In this study, the logit link function is employed to enable simultaneous probability calibration.
\end{comment}

The calibrated probabilities for the hold-out set $\mathcal{I}_{2}$ are input into the conformal prediction framework to compute the conformity scores \cite{sadinle2019least}:
\begin{equation}\label{eq:cp-score}
s_{i}(X_{i},Y_{i}) = 1-\hat{\mu}(X_{i})_{Y_{i}},
\end{equation}
where $\hat{\mu}(X_{i})_{Y_{i}}$ is the calibrated probability given the evaluation metric $X_{i}$ at $Y_{i}$, which is the ground truth label.

Next, for a given risk tolerance$\alpha$, the $(1-\alpha)^{th}$ quantile of the empirical distribution of conformity scores is computed. This quantile is then used to construct a prediction set for a new test sample 
$X_{test}$. The prediction set includes all potential labels whose conformity scores are greater than or equal to the computed quantile:
\begin{equation}\label{eq:cp-prediction}
\hat{C}_{\alpha}(X_{test}) = {y:\hat{\mu}(X_{test})\geq{1-Q_{1-\alpha}}},
\end{equation}
where $Q_{1-\alpha}$ is the $(1-\alpha)^{th}$ quantile of the conformity score distribution:
\begin{equation}\label{eq:conformal-quantile}
\centering\fontsize{9}{11}\selectfont
\frac{1}{|\mathcal{I}_{2}|+1}\sum_{i\in{\mathcal{I}_{2}}}\delta_{s_{i}}+\frac{1}{|\mathcal{I}_{2}|+1}\delta_{\infty}.
\end{equation}
Here, $s_{i} = \mathcal{S}(X_{i},Y_{i})$ and $\delta_{a}$ is a unit mass measure at $a$.
Prediction sets that either include both labels or are empty signify low confidence and high uncertainty in the predictions. The thresholds derived from the conformity scores based on the calibrated probabilities are than mapped back to the feature space to obtain thresholds on faithfulness scores.

A recent work \cite{Sudjianto2024} discusses the use of standard logistic regression for Platt scaling to map machine-generated scores to probabilities, which are then used as inputs to the conformal prediction framework. In contrast, our approach extends beyond standard logistic regression to explore more flexible methods for probability calibration. Specifically, we examine polynomial logistic regression and Generalized Additive Models (GAM), both of which are grounded in logistic regression. These efforts result in improved accuracy and greater efficiency in quantifying the uncertainty of machine-generated metrics \footnote{It is not clear what data and specific experimental set up Ref.~\cite{Sudjianto2024} used in their work, hence we have set it up from scratch with our data and our proposed overall framework to have a valid comparison.}

\subsubsection{Other Potential Approaches}
In addition, one can also resort to various other statistical tests (e.g., Kolmogorov-Smirnov test, t-test, Mann-Whitney U test, Youden's J statistic, Kullback-Leibler divergence, etc.), Bayesian methods (e.g., the threshold can be a point that minimizes the expected posterior loss), or even unsupervised clustering methods (e.g., cluster the computed scores using, for example, K-means or Gaussian mixture methods, and then identify the boundary between clusters to determine the thresholds), etc. to identify a threshold.

\section{A Concrete Example: Faithfulness}
As a specific example to demonstrate the methodology proposed in the previous section, we focus on the RAG for LLMs. RAG is a technique that enhances model responses by integrating relevant information retrieved from external knowledge sources during the generation process \cite{lewis2020retrieval}. It combines the strengths of retrieval systems and generative models to produce answers that are both contextually appropriate and grounded in up-to-date data. 

Faithfulness \cite{guo2023evaluating} is an important evaluation metric for a RAG system that measures how accurately the model's generated responses reflect the retrieved or provided source information, ensuring that outputs are reliable and trustworthy. High faithfulness prevents the dissemination of incorrect or misleading information, which is crucial for applications where accuracy is essential.

In Appendix A, we describe three of the popular open-source implementations of faithfulness (RAGAS, UpTrain and DeepEval) that will be used for our experiments.

\subsection{Data}
For the sake of reproducibility, we perform our experiments on publicly available dataset. There are few open-source datasets which contain questions, context, answers generated by LLMs, and human-annotated ground truth labels to denote if the answer is hallucinated or not. HaluBench \cite{ravi2024lynx} is a recently released and publicly available dataset comprising of around 15K real world LLM prompt and response examples arising from diverse domains (such as census data, medical research, finance, sports reports, etc.). Each entry in this dataset contains a query identifier; a passage (i.e., the context used to generate the answer); a user prompted question; an answer; and a label, which takes a value of 'PASS' or 'FAIL' depending on whether the answer is hallucinated or not.

Since most of the three hallucination scoring methodologies in our experiments hold the implicit assumption that the generated answer is a relatively long text (more than just a couple of words in the answer), we filter the data to include only those samples that have at least three tokens in the answer list. This preprocessing results in 9,616 entries in our dataset. 

Generating statements from very short answers may turn out to be a difficult task. Moreover, even in cases where the answers are long enough, sometimes RAGAS is not able to generate statements, in turn giving no score. In the experiments in the present work, we remove all the samples where RAGAS and UpTrain generated 'not a number' as the faithfulness scores thus reducing the number of samples further to 7,703 entries.

\subsection{Computational Details}
We utilize the gpt-4o-mini model as the base model to compute all the scores from all three implementations. The gpt-4o-mini\footnote{\url{https://openai.com/index/gpt-4o-mini-advancing-cost-efficient-intelligence/}} model is a lightweight version of the gpt-4o model, designed to handle a combination of text, audio, and video data, and computationally affordable for our investigation.

In RAGAS, we used the default LLM hyperparameters such as temperature and top-p. For the DeepEval framework, we specify our LLM model as gpt-4o-mini and set \verb|include_reason = True|, which includes a reason for its evaluation score. We keep other hyperparameters as default for the respective libraries. For UpTrain, we define gpt-4o-mini as our LLM model and set the evaluation metric to FACTUAL\_ACCURACY for faithfulness evaluation.

\subsection{Evaluating the Quality of Thresholds}
Each method for computing thresholds was evaluated using stratified 5-fold cross-validation with scikit-learn\footnote{\url{https://scikit-learn.org/1.5/modules/generated/sklearn.model_selection.cross_validate.html}} to ensure stability and robustness. Thresholds at various risk levels (e.g., 80\%, 90\%, 95\%, 97.5\%, and 99\%) were derived from the training datasets and tested for generalizability on test sets. Classification outcomes (hallucinated or not) were compared to ground truth labels, and evaluation metrics were computed. For conformal prediction, coverage rate and prediction set width were analyzed for accuracy and efficiency, while recall was used for other methods to assess their ability to identify true non-hallucination samples. Results were compared against baseline thresholds from standard logistic regression to highlight potential improvements in accuracy, efficiency, or informativeness. Visualizations were created to illustrate outcomes, emphasizing both strengths and limitations of the methods.

\section{Results}
Here, we describe the results from our experiments.
\subsection{The Faithfulness Scores and the Ground-truth Labels}
\begin{figure}[ht]
    \centering
    \includegraphics[width=1\linewidth]{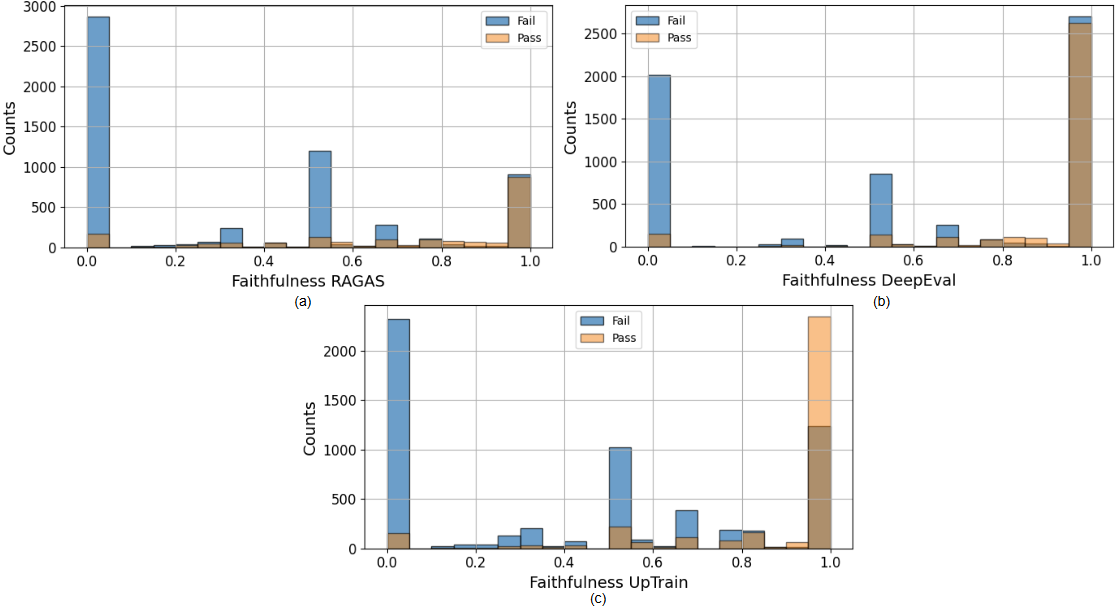}
    \caption{Conditional histograms for the faithfulness scores: (a) RAGAS; (b) DeepEval; (c) Uptrain.}
    \label{fig:histograms}
\end{figure}

Figure \ref{fig:histograms} compares the distributions of faithfulness scores from the libraries RAGAS, DeepEval, and Uptrain, respectively, while also showing their relationship with binary labels, "Pass" and "Fail." A correspondence between low scores and the "Fail" label, high score and "Pass" label would yield that the faithfulness metric is effective at detecting hallucination in the generated outputs.

The RAGAS faithfulness shows a substantial clustering of samples near the Faithfulness score being $0$, where the "Fail" label is dominant, yielding that this metric effectively assigns low scores to outputs associated with hallucination. However, there are a few peaks in between the full range of Faithfulness, and a smaller cluster exists at 1.0 where the "Pass" label is more prominent. There are also relatively few instances of "Pass" at high scores.

The DeepEval Faithfulness score exhibits large clusters at 0.0 and 1.0. The zero faithfulness score is significantly associated with the samples with the "Fail" label, confirming that low scores effectively correspond to outputs with hallucinations or factual inconsistencies. At the opposite extreme, the "Pass" label is dominant at a score of 1.0, while there is an even higher proportion still under the “Fail” category.

The UpTrain faithfulness score exhibits a more granular distribution of faithfulness scores, with clusters not only at 0.0 and 1.0 but also at intermediate values such as 0.4 and 0.6. Similar to the other libraries, the "Fail" label is heavily concentrated at a faithfulness score of 0.0, indicating the metric’s capability to detect hallucinated outputs. However, unlike DeepEval and RAGAS, UpTrain has a larger portion of “Pass” when the faithfulness score is 1.

\subsection{Thresholds}
Here, we provide results for the threshold computation using some of the methods proposed in this work. Note that as shown in Appendix C, the simplistic Z-score method (as expected) did not give useful thresholds due to the bimodal nature of the scores. Appendix D presents the results of the naive method for identifying the midpoint of bimodal distributions using histograms. Additional analyses of thresholds derived from AUC-ROC and conformal prediction are provided in Appendices E and F, respectively.

\subsection{Thresholds using KDE and Empirical Recall}
\subsubsection{Kernel Density Estimation}
The results od KDE are presented in Table \ref{tbl:comparison} and Figure \ref{fig: kde}. The KDE method performs reasonably well at lower confidence levels (e.g., 80\%), particularly for the Uptrain dataset, where the bimodal structure of the faithfulness scores is more distinct. However, as the confidence level increases, the thresholds derived from the KDE method often fail to align with the desired levels. This limitation may be attributed to the larger proportion of ambiguous scores around 0.5 in the DeepEval and RAGAS datasets, as illustrated in Figure \ref{fig: kde}. These ambiguous scores add complexity to the task of effectively separating the classes using KDE.
Furthermore, the KDE method introduces sensitivity to the choice of kernel and bandwidth parameters, which limits its robustness and adaptability across different datasets and applications.
\subsubsection{Empirical Recall}
The empirical recall was computed by iteratively varying the faithfulness score in ascending order. At lower confidence levels (e.g., 80\%), the method's output aligns well with the desired separation of classes across the libraries. However, at higher confidence levels, the thresholds derived using this method exhibit overly conservative behavior, particularly for the DeepEval dataset. The limited discriminative power of the method often results in thresholds defaulting to zero, failing to provide meaningful distinctions between classes.

\subsubsection{Thresholds using AUC-ROC}
\begin{figure}[ht]
\centering
  \includegraphics[width=\linewidth]{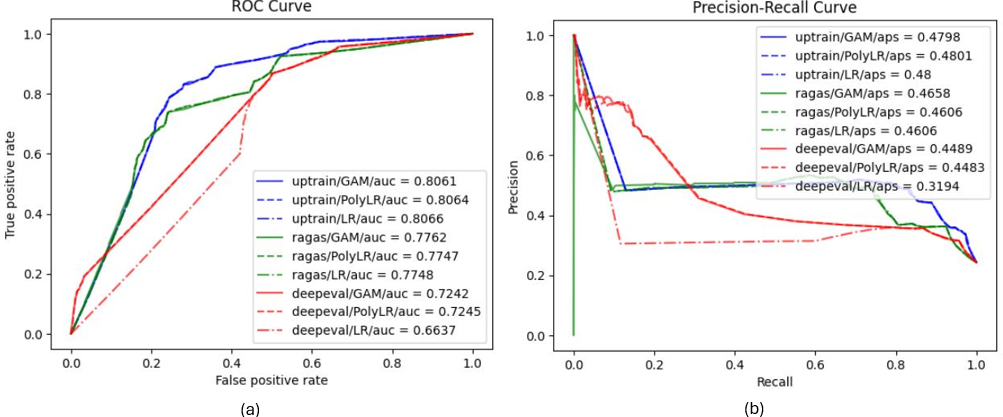}%
  \caption{Visualizations of model performance. (a) ROC curve, (b) Precision-recall curve. }\label{fig: roc-pr-curve}
\end{figure}

Figure \ref{fig: roc-pr-curve} demonstrates the ROC and precision-recall curves of various thresholding strategies. For the UpTrain and RAGAS datasets, the curves are nearly identical across all classifiers, indicating comparable performance. However, a significant improvement in performance is observed with GAM and polynomial logistic regression for the DeepEval dataset.

For datasets with higher noise levels and lower reliability, GAM and polynomial logistic regression are recommended as they better capture the nuances in relating faithfulness scores to hallucination labels. Overall, the best scores are consistently achieved either by GAM or polynomial logistic regression, highlighting their effectiveness in this context. Across the libraries, UpTrain demonstrates the highest AUC-ROC score and average precision score, followed by RAGAS and DeepEval. This observation aligns with recent work \cite{opitz2024schroedinger}, which found that AUC favors feature distributions with extreme values over more balanced ones. As shown in the histogram of DeepEval, the extreme effect is less pronounced compared to the other two datasets.

\subsection{Thresholds using Conformal Prediction}

Table \ref{tbl:cp-performance} compares the conformal prediction performance across different underlying classifiers at confidence levels of 80\%, 90\%, 95\%, 97.5\%, and 99\%. All methods achieve coverage rates that converge to the specified confidence levels, validating the conformal prediction framework. Another critical aspect of comparison is the size of the prediction sets, which directly measures how informative the predictions are. Counting wins and losses, GAM consistently produces narrower prediction sets overall, followed by polynomial logistic regression. Since no empty prediction sets were observed at the examined confidence levels, narrower prediction sets indicate that GAM and polynomial logistic regression result in fewer cases where both labels are included in the prediction set. This suggests that these methods offer higher efficiency, greater informativeness, and improved accuracy compared to the baseline approach.

\begin{table*}[ht]
\centering
{\small
\begin{tabular}{lr|rrr|rrr|rrr}
\hline
                      & \multicolumn{1}{l|}{}                 & \multicolumn{3}{l|}{GAM}                                                                     & \multicolumn{3}{l|}{Polynomial logistic regression}                                          & \multicolumn{3}{l}{Standard logistic regression}                                                                \\ \hline
\multicolumn{1}{l|}{Library}  & \multicolumn{1}{l|}{1-$\alpha$} & \multicolumn{1}{l}{threshold} & \multicolumn{1}{l}{coverage} & \multicolumn{1}{l|}{width} & \multicolumn{1}{l}{threshold} & \multicolumn{1}{l}{coverage} & \multicolumn{1}{l|}{width} & \multicolumn{1}{l}{threshold} & \multicolumn{1}{l}{coverage} & \multicolumn{1}{l}{width} \\ \hline
\multicolumn{1}{l|}{Uptrain}         & 80\%                                  & 0.54                             & 0.8085                       & 1.1485                     & 0.597                            & 0.8038                       & 1.1363                     & 0.5589                           & 0.8085                       & 1.1485                    \\
\multicolumn{1}{l|}{}         & 90\%                                  & 0.451                            & 0.9835                       & 1.627                      & 0.3802                           & 0.9865                       & 1.64                       & 0.3722                           & 0.9869                       & 1.6408                    \\
\multicolumn{1}{l|}{}         & 95\%                                  & 0.451                            & 0.9835                       & 1.627                      & 0.3802                           & 0.9865                       & 1.64                       & 0.3722                           & 0.9869                       & 1.6408                    \\
\multicolumn{1}{l|}{}         & 97.5\%                                & 0.451                            & 0.9835                       & 1.627                      & 0.3802                           & 0.9865                       & 1.64                       & 0.3722                           & 0.9869                       & 1.6408                    \\
\multicolumn{1}{l|}{}         & 99\%                                  & 0.2555                           & 0.991                        & 1.6796                     & 0.29                             & 0.991                        & 1.6796                     & 0.29                             & 0.991                        & 1.6796                    \\ \hline
\multicolumn{1}{l|}{DeepEval} & 80\%                                  & 0.4074                           & 0.9551                       & 1.6374                     & 0.3894                           & 0.9559                       & 1.6391                     & 0.4374                           & 0.9844                       & 1.7144                    \\
\multicolumn{1}{l|}{}         & 90\%                                  & 0.4074                           & 0.9551                       & 1.6374                     & 0.3894                           & 0.9559                       & 1.6391                     & 0.4374                           & 0.9844                       & 1.7144                    \\
\multicolumn{1}{l|}{}         & 95\%                                  & 0.4074                           & 0.9608                       & 1.6455                     & 0.3878                           & 0.9616                       & 1.6472                     & 0.4374                           & 0.9844                       & 1.7144                    \\
\multicolumn{1}{l|}{}         & 97.5\%                                & 0.3939                           & 0.9749                       & 1.6795                     & 0.3598                           & 0.9755                       & 1.6803                     & 0.4374                           & 0.9844                       & 1.7144                    \\
\multicolumn{1}{l|}{}         & 99\%                                  & 0.1419                           & 0.9938                       & 1.8444                     & 0.14                             & 0.9938                       & 1.8444                     & 0.14                             & 0.9938                       & 1.8444                    \\ \hline
\multicolumn{1}{l|}{RAGAS}    & 80\%                                  & 0.3988                           & 0.8157                       & 1.243                      & 0.3916                           & 0.8183                       & 1.2477                     & 0.4258                           & 0.8142                       & 1.24                      \\
\multicolumn{1}{l|}{}         & 90\%                                  & 0.0755                           & 0.9814                       & 1.6172                     & 0.1138                           & 0.9814                       & 1.6187                     & 0.0925                           & 0.9817                       & 1.6195                    \\
\multicolumn{1}{l|}{}         & 95\%                                  & 0.0755                           & 0.9814                       & 1.6172                     & 0.1138                           & 0.9814                       & 1.6187                     & 0.0925                           & 0.9817                       & 1.6195                    \\
\multicolumn{1}{l|}{}         & 97.5\%                                & 0.0755                           & 0.9814                       & 1.6172                     & 0.1138                           & 0.9814                       & 1.6187                     & 0.0925                           & 0.9817                       & 1.6195                    \\
\multicolumn{1}{l|}{}         & 99\%                                  & 0                                & 1                            & 2                          & 0                                & 1                            & 2                          & 0                                & 1                            & 2                         \\ \hline
\end{tabular}
}
\caption{Conformal prediction performance at different confidence levels.}\label{tbl:cp-performance}
%\vspace{-8mm}
\end{table*}

\subsection{Comparison across different methods}
Tables \ref{tbl:cp-performance} and \ref{tbl:comparison}  present the results of different methods across various confidence levels, including 80\%, 90\%, 95\%, 97.5\%, and 99\%, based on a stratified 5-fold cross-validation strategy. The reported results include thresholds in terms of faithfulness scores and performance metrics (coverage rate for conformal prediction, and recall for other methods). 

Overall, all methods achieve valid coverage aligned with the pre-specified confidence levels. A closer examination of Table \ref{tbl:comparison} reveals that KDE provides more precise estimations at lower confidence levels (e.g., 80\%). However, as the confidence level increases, thresholds derived from these approaches often default to zero faithfulness scores, resulting in 100\% recall values that fail to provide meaningful insights. This highlights the limited capability of these methods in effectively identifying thresholds, particularly when the confidence requirement is high.

Additionally, the precision-recall curve tends to produce the most conservative thresholds, leading to higher recall values but at the cost of informativeness. In contrast, as shown in Table \ref{tbl:cp-performance}, conformal prediction demonstrates better discriminative power in identifying thresholds at various confidence levels, maintaining guaranteed coverage despite the limited dispersion in faithfulness scores.

Here, the risk levels measured in the two tables \ref{tbl:cp-performance} and \ref{tbl:comparison} differ slightly—coverage rate for conformal prediction versus recall for precision-recall curves—making the corresponding thresholds not directly comparable.

\section{The Conclusions}

We addressed the critical challenge of determining appropriate thresholds for evaluation metrics in LLMs. Recognizing that threshold selection significantly impacts the reliability and societal implications of LLM deployments, we proposed a systematic and statistically rigorous methodology that integrates stakeholder risk preferences into the threshold determination process. By drawing parallels with financial risk assessment techniques, such as prospect theory, we proposed to quantify stakeholders' risk aversion and translate it into statistical confidence levels, ensuring that threshold selection is both statistically sound and tailored to the specific needs of the deploying entities.

We explored various statistical methods for identifying optimal thresholds, including Receiver Operating Characteristic (ROC) curves, kernel density estimation (KDE), and conformal predictions. Each method offers unique advantages in quantifying the association between evaluation metric scores and ground truth labels, enabling practitioners to select thresholds that maximize model performance while adhering to acceptable error rates.

Our findings underscore the importance of a comprehensive approach that combines statistical rigor with stakeholder engagement. By systematically assessing risk tolerance and applying appropriate statistical techniques, developers can set thresholds that enhance the accuracy and reliability of LLM outputs while mitigating potential harms associated with incorrect or misleading responses.

We emphasize that the procedure proposed in the present work can also be extended to identify thresholds for evaluation metrics for multi-model AI systems in the future.

\begin{table*}[]
\centering
{\small
\begin{tabular}{lr|rr|rr|rr|rr|rr}
\hline
Classifier                    & \multicolumn{1}{l|}{}                 & \multicolumn{1}{l}{GAM}          & \multicolumn{1}{l|}{}       & \multicolumn{1}{l}{Polynomial LR}         & \multicolumn{1}{l|}{}       & \multicolumn{1}{l}{Standard LR}     & \multicolumn{1}{l|}{}       & \multicolumn{1}{l}{-}            & \multicolumn{1}{l|}{}       & \multicolumn{1}{l}{-}            & \multicolumn{1}{l}{}       \\ \hline
method                        & \multicolumn{1}{l|}{}                 & \multicolumn{2}{l|}{precision-recall curve}                    & \multicolumn{2}{l|}{precision-recall curve}                    & \multicolumn{2}{l|}{precision-recall curve}                    & \multicolumn{2}{l|}{empirical   recall}                        & \multicolumn{1}{l}{KDE}          & \multicolumn{1}{l}{}       \\ \hline
\multicolumn{1}{l|}{Library}  & \multicolumn{1}{l|}{1-$\alpha$} & \multicolumn{1}{l}{threshold} & \multicolumn{1}{l|}{recall} & \multicolumn{1}{l}{threshold} & \multicolumn{1}{l|}{recall} & \multicolumn{1}{l}{threshold} & \multicolumn{1}{l|}{recall} & \multicolumn{1}{l}{threshold} & \multicolumn{1}{l|}{recall} & \multicolumn{1}{l}{threshold} & \multicolumn{1}{l}{recall} \\ \hline
\multicolumn{1}{l|}{Uptrain}  & 80\%                                  & 0.3514                           & 0.9358                      & 0.4096                           & 0.9358                      & 0.3648                           & 0.9358                      & 0.7500                           & 0.8164                      & 0.6555                           & 0.8404                     \\
\multicolumn{1}{l|}{}         & 90\%                                  & 0.2375                           & 0.9626                      & 0.2839                           & 0.9652                      & 0.2424                           & 0.9652                      & 0.5000                           & 0.9322                      & 0.2569                           & 0.9648                     \\
\multicolumn{1}{l|}{}         & 95\%                                  & 0.1923                           & 0.9679                      & 0.2293                           & 0.9679                      & 0.1869                           & 0.9679                      & 0.3333                           & 0.9594                      & 0.1255                           & 0.9717                     \\
\multicolumn{1}{l|}{}         & 97.5\%                                & 0.1312                           & 1.0000                      & 0.1667                           & 1.0000                      & 0.1333                           & 1.0000                      & 0.0000                           & 1.0000                      & 0.0721                           & 0.9733                     \\
\multicolumn{1}{l|}{}         & 99\%                                  & 0.1248                           & 1.0000                      & 0.0979                           & 1.0000                      & 0.0667                           & 1.0000                      & 0.0000                           & 1.0000                      & 0.0000                           & 1.0000                     \\ \hline
\multicolumn{1}{l|}{DeepEval} & 80\%                                  & 0.4820                           & 0.9305                      & 0.4856                           & 0.9305                      & 0.4053                           & 0.9332                      & 0.4864                           & 0.8052                      & 0.1662                           & 0.9562                     \\
\multicolumn{1}{l|}{}         & 90\%                                  & 0.3876                           & 0.9412                      & 0.3857                           & 0.9412                      & 0.2839                           & 0.9412                      & 0.2017                           & 0.9092                      & 0.1459                           & 0.9567                     \\
\multicolumn{1}{l|}{}         & 95\%                                  & 0.3662                           & 0.9412                      & 0.3627                           & 0.9412                      & 0.2036                           & 0.9492                      & 0.0000                           & 1.0000                      & 0.1209                           & 0.9567                     \\
\multicolumn{1}{l|}{}         & 97.5\%                                & 0.0000                           & 1.0000                      & 0.0000                           & 1.0000                      & 0.0000                           & 1.0000                      & 0.0000                           & 1.0000                      & 0.0000                           & 1.0000                     \\
\multicolumn{1}{l|}{}         & 99\%                                  & 0.0000                           & 1.0000                      & 0.0000                           & 1.0000                      & 0.0000                           & 1.0000                      & 0.0000                           & 1.0000                      & 0.0000                           & 1.0000                     \\ \hline
\multicolumn{1}{l|}{RAGAS}    & 80\%                                  & 0.3108                           & 0.8690                      & 0.2645                           & 0.8690                      & 0.2640                           & 0.8690                      & 0.6667                           & 0.8500                      & 0.1297                           & 0.9215                     \\
\multicolumn{1}{l|}{}         & 90\%                                  & 0.2296                           & 0.9118                      & 0.1509                           & 0.9118                      & 0.1465                           & 0.9118                      & 0.5000                           & 0.9359                      & 0.1069                           & 0.9247                     \\
\multicolumn{1}{l|}{}         & 95\%                                  & 0.4982                           & 1.0000                      & 0.0660                           & 1.0000                      & 0.0290                           & 1.0000                      & 0.3071                           & 0.9514                      & 0.0420                           & 0.9354                     \\
\multicolumn{1}{l|}{}         & 97.5\%                                & 0.0000                           & 1.0000                      & 0.0000                           & 1.0000                      & 0.0000                           & 1.0000                      & 0.0000                           & 1.0000                      & 0.0000                           & 1.0000                     \\
\multicolumn{1}{l|}{}         & 99\%                                  & 0.0000                           & 1.0000                      & 0.0000                           & 1.0000                      & 0.0000                           & 1.0000                      & 0.0000                           & 1.0000                      & 0.0000                           & 1.0000                     \\ \hline
\end{tabular}
}
\caption{Comparison across different methods at various confidence levels.}\label{tbl:comparison}
%\vspace{-8mm}
\end{table*}

\section*{Acknowledgement}
The views expressed here are those of the authors alone and not of BlackRock, Inc.

\bibliographystyle{abbrv}
\bibliography{refs}

\clearpage
\appendix

\section{Appendix A: Faithfulness Implementations}\label{app:libraries}
In this Appendix, we provide details of various faithfulness implementations used in the experiments in this study.

\noindent\textbf{RAGAS Faithfulness Metric:}
The Faithfulness metric as proposed in Ref.~\cite{es2023ragas} and implemented in the RAGAS\footnote{\url{https://github.com/explodinggradients/ragas}} library evaluates how faithful the LLM generated answer is to a retrieved context. Using another LLM, this evaluation metric amounts to generate claims made in the response and determines if each claim can be inferred from the context, producing a score scaled between 0 and 1, where higher scores indicate better alignment of the answer to the given context. This metric is particularly useful in detecting inaccuracies or unsupported claims in generated outputs. The process involves breaking the answer into statements, then cross-checking each statement against the context, and finally calculating a faithfulness score based on the proportion of truthful claims. RAGAS also integrates Vectara's HHEM-2.1-Open\footnote{\url{https://huggingface.co/vectara/hallucination_evaluation_model}}, an open-source classifier trained to detect hallucinations, which can enhance faithfulness evaluation in production environments. 

\noindent\textbf{DeepEval's Faithfulness:}
DeepEval's Faithfulness\footnote{\url{https://docs.confident-ai.com/docs/metrics-hallucination}} metric evaluates whether an LLM generated output factually aligns with a given retrieved context. Unlike the RAGAS Faithfulness, this measure emphasizes detecting contradictions between the actual output and the retrieved context. The metric quantifies faithfulness by calculating the proportion of truthful claims in the output relative to the total claims made, with each claim assessed against the retrieval context for factual consistency. DeepEval also offers self-explaining evaluations, providing reasons for the assigned scores. This metric also suffers from the few limitations that RAG Faithfulness does, namely, inability to generating statements from short answers, though in our experiments we did not find any example without a finite score when used this library.

\noindent\textbf{UpTrain's Factual Accuracy Metric:}
UpTrain's factual accuracy evaluation\footnote{\url{https://docs.uptrain.ai/predefined-evaluations/context-awareness/factual-accuracy}} metric measures the degree to which the claims made in the generated output align with the retrieved context, i.e., first, it splits the generated answer to individual claims, then an LLM evaluates whether each of the claims is correct on the basis of the given context. A difference here is that the correctness is not evaluated in a binary fashion, rather they are labeled as Completely Right (score 1), Completely Wrong (score 0) and Ambiguous (score 0.5). Finally, the mean of the scores over individual claims is provided as the final score.

\section{Appendix B: Results of Statistical Tests}
In this Section, we provide results of statistical tests for the outputs from the three libraries.

\noindent\textbf{t-test:} The t-test is a statistical hypothesis test used to determine whether there is a significant difference between the means of two groups. We used it to examine whether there is a significant difference in the faithfulness scores between the 'FAIL' and 'PASS' labels. 

We computed the independent t-test to determine whether there is a significant difference between the faithfulness scores of the 'Fail' and 'Pass' labels, and the results are shown in table \ref{tab:t_test}. These results indicate there is a significant difference between the scores of the 'FAIL' and 'PASS' labels for all libraries. 
\begin{table}
    \centering
    \begin{tabular}{l|l|l}
\hline
Library  & t-statistic & p-value                 \\ \hline
RAGAS    & 41.78       & 0                       \\
DeepEval & 27.48       & $1.09 \times 10^{-158}$ \\
Uptrain  & 46.47       & 0                       \\ \hline
\end{tabular}
    \caption{ Independent samples t-test result}
    \label{tab:t_test}
\end{table}

\noindent\textbf{Mann-Whitney Test:} Besides the t-test, we also did a Mann-Whitney U test, which is another non-parametric statistical hypothesis test used to determine if there is a significant difference between the distributions of two independent groups. By applying the Mann-Whitney U test, we can assess whether the distributions of the faithfulness scores differ significantly between the 'Fail' and 'Pass' labels. The results are shown in table \ref{tab:u_test}. These results indicate a significant difference between the scores of the 'FAIL' and 'PASS' labels for all libraries.

\begin{table}
    \centering
    \begin{tabular}{l|l|l}
\hline
Library  & U-statistic & p-value                 \\ \hline
RAGAS    & 2438046.5   & $4.45 \times 10^{-307}$ \\
DeepEval & 3648584.5   & $1.20 \times 10^{-117}$ \\
Uptrain  & 2076793.0   & 0                       \\ \hline
\end{tabular}
    \caption{Mann-Whitney U Test Result}
    \label{tab:u_test}
\end{table}

\section{Appendix C: Thresholds using Z-score}
We used a 95\% confidence level to calculate the threshold using the Z-score. Table \ref{tab:z_score} shows the mean and standard deviation of the scores from each library for the complete data. By using the formula\ref{eq:z-score} and a Z-score of 1.96, we calculated the lower and upper bounds for the thresholds. Since all lower thresholds are below 0 and all upper thresholds are above 1, of course, this intuitive and straightforward approach does not make sense. 
\begin{table}[h]
    \centering
    \begin{tabular}{l|l|l|l|l}
\hline
Library  & Mean & Std Dev & LB    & UB   \\ \hline
RAGAS    & 0.44 & 0.40    & -0.35 & 1.22 \\
DeepEval & 0.63 & 0.42    & -0.19 & 1.45 \\
Uptrain  & 0.54 & 0.41    & -0.27 & 1.34 \\ \hline
\end{tabular}
    \caption{Z-score Thresholds using 95\% confidence level. LB and UB refer to Lower Bound and Upper Bound.}
    \label{tab:z_score}
\end{table}

\section{Appendix D: Mid Point Identification through Histograms \label{appendix}}
\begin{figure*}[ht]
\centering
  \includegraphics[width=\linewidth]{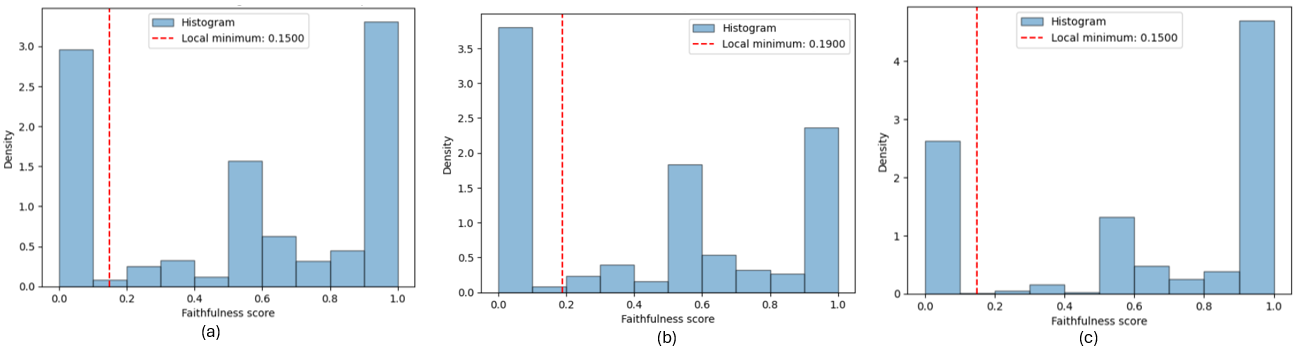}%
  \caption{Example visualizations of thresholds identified using local minimum. (a) UpTrain, (b) RAGAS, (c) DeepEval.}\label{fig: local-min}
\end{figure*}
In Figure \ref{fig: local-min}, the corresponding recall values are 0.97, 0.87, and 0.96 for Uptrain, DeepEval, and RAGAS, respectively. These relatively high recall scores confirm the method's effectiveness in identifying the midpoint for separating the bimodal distribution. However, associating the midpoints with specific confidence levels is challenging, which limits the method's adaptability and applicability for determining thresholds.

\begin{figure*}[ht]
\centering
  \includegraphics[width=\linewidth]{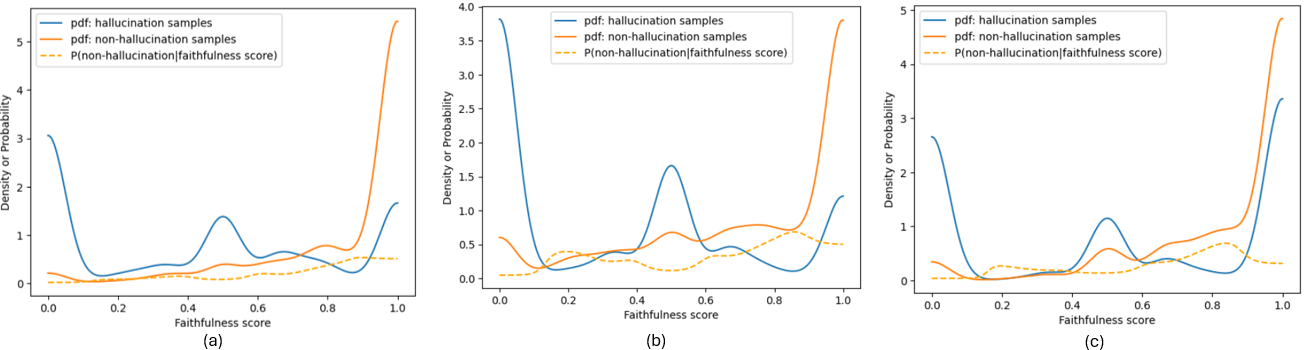}%
  \caption{Example visualizations of thresholds identified using KDE. (a) UpTrain, (b) RAGAS, (c) DeepEval.}\label{fig: kde}
\end{figure*}
\begin{figure*}[ht]
\centering
  \includegraphics[width=\linewidth]{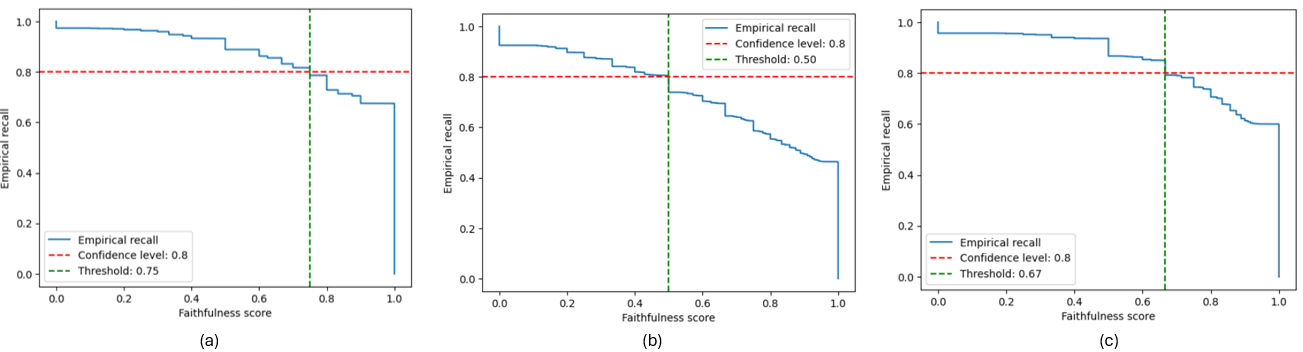}%
  \caption{Visualizations of thresholds identified using empirical recall curve. (a) UpTrain, (b) RAGAS, (c) DeepEval.}\label{fig: erecall}
\end{figure*}

\section{Appendix E: Further Analysis of Thresholds from AUC-ROC}
\begin{figure*}[h]
\centering
  \includegraphics[width=\linewidth]{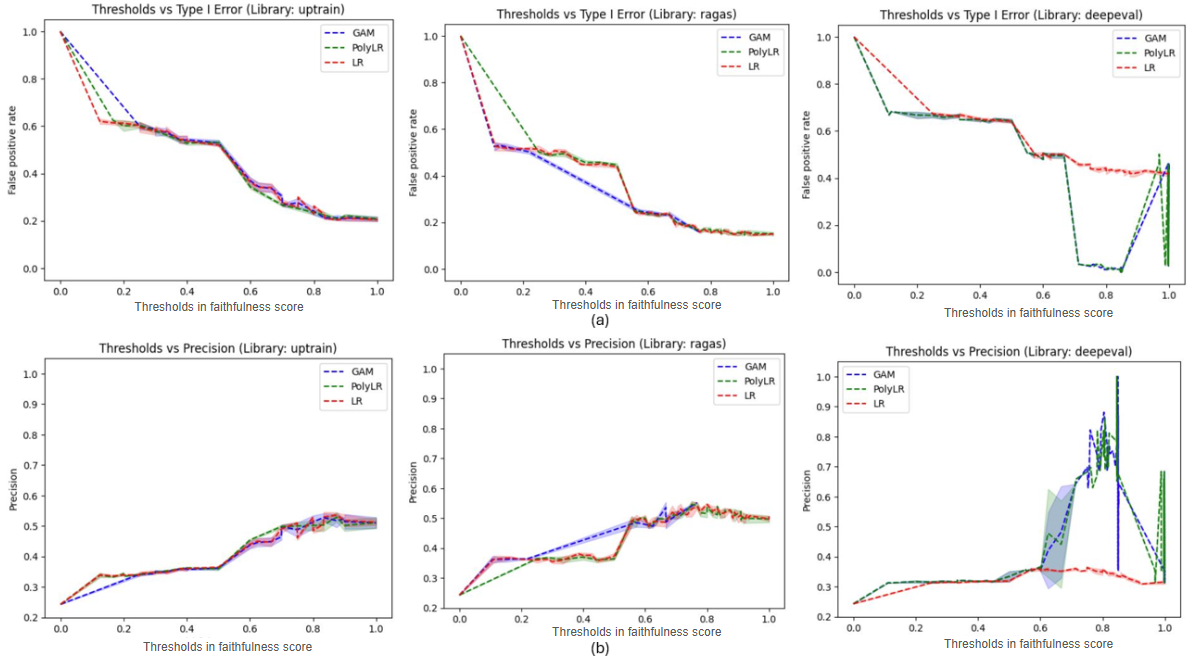}%
  \caption{Thresholds at different risk levels. (a): Thresholds vs type I error, (b): Thresholds vs precision.}\label{fig: roc-pr-thresholds}
\end{figure*}

Further examination on Figure \ref{fig: roc-pr-curve} reveals that, for the Uptrain and RAGAS datasets, GAM and polynomial logistic regression produce thresholds that are nearly identical to those derived from standard logistic regression across various risk levels. However, for the DeepEval dataset, the two approaches yield noticeably different thresholds, particularly at higher faithfulness scores. This divergence can be attributed to the increased confusion in differentiating hallucinations as the faithfulness score approaches 1, as observed in the plots in Figures \ref{fig:histograms}. GAM and polynomial logistic regression effectively captures this behavior: as the faithfulness score increases, it initially lowers the false positive rate but subsequently raises it when faithfulness is closer to 1. A similar pattern is observed for precision, where it first improves, then declines as the faithfulness score nears 1. This dynamic response highlights the ability of polynomial logistic regression to adapt to nuanced patterns in the DeepEval scores. 
Figure \ref{fig: roc-pr-thresholds} further illustrates the identified thresholds across different risk tolerances, measured by the false positive rate (type I error) and precision. The mean and standard deviation of the thresholds obtained from five-fold cross-validation are plotted against various risk levels. Intuitively, as the thresholds increase from zero to one, cases with faithfulness scores exceeding the thresholds are classified as positive (passing the hallucination test), resulting in a decreasing false positive rate and an increasing precision. A closer examination of the results from standard logistic regression on the DeepEval dataset reveals that as the faithfulness score progresses from zero to one, the corresponding changes in false positive rate and precision are the smallest. This indicates that the hallucination outcomes are the least sensitive to the faithfulness scores generated by the DeepEval library compared to those from Uptrain and RAGAS.

\section{Appendix F: Further Analysis of Thresholds from Conformal Prediction}
\begin{figure*}[ht]
\centering
  \includegraphics[width=\linewidth]{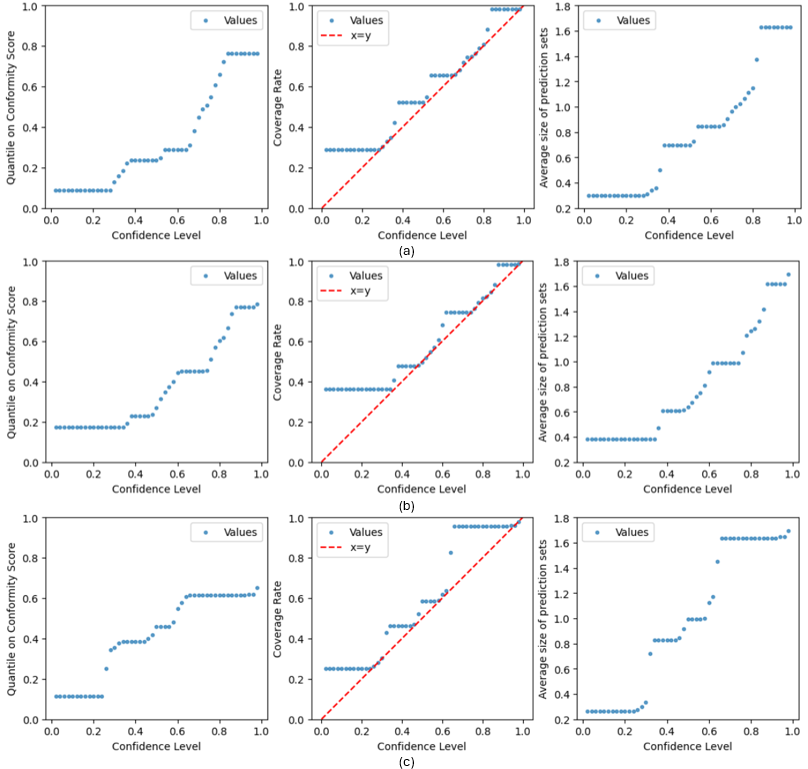}%
  \caption{Visualizations of conformal prediction performance at various confidence levels. (a) UpTrain, (b) RAGAS, (c) DeepEval}\label{fig:cp-performance}
\end{figure*}

The performance of conformal prediction was evaluated at various confidence levels using standard logistic regression, polynomial logistic regression and GAMs as the underlying classifiers as shown in Figure \ref{fig:cp-performance}: with confidence levels ranging from 0.02 to 1 in increments of 0.02. Ideally, the coverage rate versus confidence level should align with the 45-degree line. However, it is observed that, in most cases, the method achieves coverage rates higher than the target levels (as shown in the second figure). While this validates the conformal predictor, it also indicates reduced efficiency, as prediction sets tend to be wider (illustrated in the third figure). This inefficiency may be attributed to limited dispersion in the quantiles.

\begin{figure}[ht]
\centering
  \includegraphics[width=\linewidth]{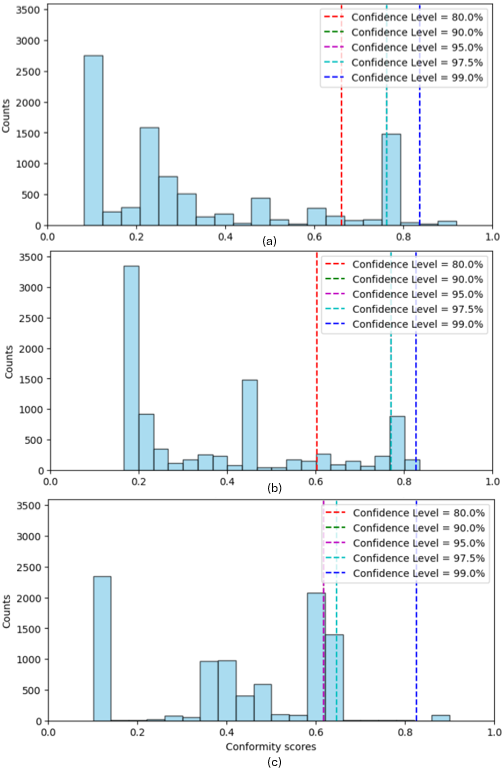}%
  \caption{Distribution of conformity scores with thresholds. (a) UpTrain, (b) RAGAS, (c) DeepEval.}\label{fig:cp-thresholds}
\end{figure}

The distribution of conformity scores, along with empirical quantiles at 80\%, 90\%, 95\%, 97.5\%, and 99\% confidence levels, is shown in Figure \ref{fig:cp-thresholds}. Notably, for the DeepEval dataset, a significant proportion of scores concentrate around 0.5, indicating a 50\% probability for binary classification. This high level of ambiguity can likely be attributed to the challenge of distinguishing between pass and fail outcomes when the faithfulness score equals 1, as was already observed in Figure \ref{fig:histograms}.

\end{document}